\documentclass[letterpaper, 10 pt, conference]{ieeeconf}
\usepackage[english]{babel}
\usepackage[utf8]{inputenc}
\usepackage{amsmath}
\usepackage{breqn}
\usepackage{filecontents,lipsum}
\usepackage[noadjust]{cite}
\usepackage{cite}
\usepackage{subcaption}
\usepackage{graphicx}
\usepackage[version=4]{mhchem}
\usepackage{siunitx}
\usepackage{longtable,tabularx}
\usepackage{caption}
\usepackage{float}
\usepackage{booktabs}
\usepackage{authblk}
\usepackage{amsfonts}
\usepackage{caption}
\usepackage{algorithm}
\usepackage{algpseudocode}
\usepackage{siunitx}
\usepackage{setspace}
\usepackage[export]{adjustbox}
\usepackage{multirow}
\usepackage{balance}
\usepackage{microtype}
\def\p(#1|#2){p(#1\,|\,#2)}
\def\q(#1|#2){q(#1\,|\,#2)}

\usepackage{xcolor}

\newcommand{\norm}[1]{\left\lVert#1\right\rVert}

\newtheorem{thm}{Theorem}

\renewcommand{\thetable}{\arabic{table}}
\algnewcommand{\Inputs}[1]{%
  \State \textbf{Inputs:} 
   \hspace*{0.3em}\parbox[t]{\linewidth}{\raggedright #1}
}
\algnewcommand{\Initialize}[1]{%
  \State \textbf{Initialize:}
   \hspace*{0.3em}\parbox[t]{.8\linewidth}{\raggedright #1}
}
\algnewcommand{\Output}[1]{%
  \State \textbf{Outputs:}
   \hspace*{0.3em}\parbox[t]{.8\linewidth}{\raggedright #1}
}
\title{\bf{Sampling-Based Nonlinear MPC of Neural Network Dynamics with Application to Autonomous Vehicle Motion Planning}
}
\IEEEoverridecommandlockouts
\author[1]{Iman Askari\textsuperscript{1}, Babak Badnava\textsuperscript{1}, Thomas Woodruff\textsuperscript{2}, Shen Zeng\textsuperscript{3} and Huazhen Fang\textsuperscript{1}
\thanks{\textsuperscript{1}I. Askari, B. Badnava and H. Fang are with the Department of Mechanical Engineering, University of
Kansas, Lawrence, KS 66045, USA (e-mail: {\tt\small askari, babak.badnava, fang@ku.edu}).}
\thanks{\textsuperscript{2}T. Woodruff is with the Department of Electrical Engineering and Computer Science, University of
Kansas, Lawrence, KS 66045, USA (e-mail: {\tt\small tjwoodruff@ku.edu}).}
\thanks{\textsuperscript{3}S. Zeng is with the Department of Electrical and System Engineering, Washington University, St. Louis, MO 63130, USA (e-mail: {\tt\small s.zeng@wustl.edu}) and was supported by the NSF grant CMMI-1933976.}

}
\begin{document}

\maketitle
\begin{abstract} 
Control of machine learning models has emerged as an important paradigm for a broad range of robotics applications. In this paper, we present a sampling-based nonlinear model predictive control (NMPC) approach for control  of neural network dynamics. We show its design in two parts: 1)  formulating conventional optimization-based NMPC as a Bayesian state estimation problem, and 2) using particle filtering/smoothing to achieve the estimation. Through a principled sampling-based implementation, this approach can potentially make effective searches in the control action space for optimal control  and also facilitate   computation toward  overcoming the challenges caused by neural network dynamics. 
We   apply  the proposed NMPC approach to motion planning for autonomous vehicles. The specific problem considers nonlinear unknown vehicle dynamics modeled as neural networks as well as dynamic on-road driving scenarios. The approach shows significant effectiveness in successful motion planning in case studies. 
\end{abstract}

\section{Introduction} 
\label{introduction}

Machine learning has risen as an important way for modeling and control of complex dynamic systems to accelerate challenging robotics applications~\cite{Nguyen-Tuong:CP:2011,Jordan:Science:2015,Pierson:AR:2017}. This emerging field has witnessed advances, especially based on two frameworks. The first framework exploits reinforcement learning to train robots such that they learn optimal control policies from experiences to accomplish certain goals~\cite{Kober:IJRR:2013}. Despite successes in tackling various tough problems, this framework in general demands hefty amounts of training data and offers difficult generalization to tasks different from specified goals~\cite{Kober:IJRR:2013}. By contrast, the second framework seeks to integrate machine learning-based modeling with optimal control~\cite{Nguyen-Tuong:CP:2011}. Specifically, it extracts data-driven models for robotic systems and then synthesizes optimal control using such explicit models. The corresponding robot control methods would arguably present higher data efficiency and generalizability. Associated with this, a still open question is how to enable control design that well fits with machine learning models, which continually calls for the development of constructive methods.

In this study, we consider the problem of controlling neural network models for dynamic systems and investigate nonlinear model predictive control (NMPC) for them. Neural networks have gained increasing use in modeling robots with complicated dynamics. However, the application of NMPC to them is non-trivial. Conventional NMPC entails online constrained optimization, but numerical optimization of neural network dynamics can be extremely burdensome due to a mix of nonlinearity, nonconvexity, and heavy computation. This is particularly true when a neural network has many hidden layers and nodes. In addition, some robotics applications involve cost functions that have zero gradients, precluding the use of gradient-based optimization.

To address the challenge, we propose to leverage a new NMPC approach, which was developed in our previous study~\cite{Askari:ACC:2021} and named {\em constraint-aware particle filtering/smoothing NMPC} (CAP-NMPC), to control neural network dynamics. Departing from the optimization-based view, the CAP-NMPC interprets NMPC through the lens of Bayesian estimation and designs a constrained particle filtering/smoothing method to achieve it. This approach, at its core, uses sequential Monte Carlo sampling to estimate the optimal control actions from a reference signal that is to be followed by the neural network model over a receding horizon. The sampling-based implementation is arguably advantageous in several ways for control of neural network dynamics. It can make sufficient search for the best control actions across a large control space when using adequate numbers of particles.
The sampling-based computation also obviates the need for iteratively computing gradients and is relatively easy and efficient to achieve.

We further consider the problem of motion planning for autonomous vehicles. We construct a neural network model to capture the vehicle dynamics and then use the CAP-NMPC to perform motion planning in dynamic driving scenarios (e.g., lane changing and moving obstacles). The simulation results demonstrate the effectiveness of the design.

\section{Related Work}
\label{literature}

Neural networks have proven useful for data-driven modeling of many systems that resist analytical modeling or suffer significant modeling uncertainty.
A common approach is using feedforward neural networks to approximate the state dynamics functions. While simple network structures can be enough for some systems~\cite{Draeger:CSM:1995,Piche:NIPS:1999,Williams:ICRA:2017, Williams:TRO:2018}, the recent development of deep neural networks has allowed to capture more complex dynamics~\cite{Broad:arXiv:2018,Nagabandi:ICRA:2018}. Further, recurrent neural networks can effectively learn closed-loop or residual dynamics~\cite{Garimella:2018,Rankovic:IJCCC:2012}.

	Based on the prediction by a neural network model, NMPC can be readily designed to predictively optimize a system's behavior, as is pursued in~\cite{Draeger:CSM:1995,Piche:NIPS:1999,Williams:ICRA:2017, Williams:TRO:2018,Broad:arXiv:2018,Nagabandi:ICRA:2018,Garimella:2018,Rankovic:IJCCC:2012}. Yet, numerical optimization in this setting is recognized as a thorny issue~\cite{Williams:TRO:2018,Broad:arXiv:2018}, as gradient-based methods often find themselves inadequate and computationally expensive in the face of nonconvex optimization of neural network dynamics. Gradient-free NMPC is thus desired, and the sampling-based approach holds significant promise. A simple random-sampling shooting method is used in~\cite{Nagabandi:ICRA:2018} to treat NMPC of a deep neural network model, which generates control action sequences randomly and chooses the sequence that leads to the highest expected cumulative reward. The study in~\cite{Williams:TRO:2018} formulates information-theoretic NMPC for neural network models and then synthesizes an algorithm based on iteratively weighting sampled sequences of control actions. A direct estimation approach to motion planning for autonomous vehicles is presented in~\cite{Berntorp:TIV2019}, where the planning requirements are modeled as measurements and the trajectory of the vehicle is inferred using a particle filter. The CAP-NMPC approach differs from~\cite{Nagabandi:ICRA:2018,Williams:TRO:2018, Berntorp:TIV2019} on three aspects. First, it is developed from the perspective of Bayesian estimation and builds on particle filtering/smoothing to solve the NMPC problem, which is different that the direct filtering approach in~\cite{ Berntorp:TIV2019}. This feature also allows for diverse realizations of the CAP-NMPC since the literature includes a rich set of particle filtering/smoothing techniques. Second, rather than randomly creating all the control action samples at once and then evaluating their competence (e.g., cumulative rewards or weights), this approach generates samples sequentially based on the neural network model and then weights and resamples them. This principled manner has the potential to search the control space more effectively. Third, by design, the CAP-NMPC approach takes generic state and input constraints into account, as needed for robotic systems subject to operating constraints.

A critical function of autonomous driving, motion planning has attracted enormous interest in the past decade. State of the art has three main approaches: input space discretization with collision checking, randomized planning, and NMPC~\cite{Schwarting:ARCRAS:2018,Gonzalez:TITS:2016,Paden:TIV:2016}. Among them, NMPC can optimize vehicle motion using sophisticated vehicle models under practical constraints, thus capable of planning and executing safety-critical, aggressive (near-limit), or emergency maneuvers in complex environmental conditions~\cite{Paden:TIV:2016}. NMPC-based motion planning has gained a growing body of work recently, e.g.,~\cite{Williams:TRO:2018,Chen:TIV:2019,Brunner:CDC:2017,Nolte:IVS:2017,Cardoso:ICRA:2017,Liu:IVS:2017,Gao:DSCC:2010}, just to highlight a few. However, almost all studies, except~\cite{Williams:TRO:2018}, consider physics-based vehicle models to our knowledge, even though
modeling errors or biases are inevitable in the real world. Neural network vehicle models have demonstrated substantial achievements recently~\cite{Spielberg:SR:2019}. This paper will study motion planning based on them, using the CAP-NMPC approach as the enabling tool.

\section{The {CAP-NMPC} Approach}

In this section, we provide an overview of the CAP-NMPC approach. A more detailed description is available in~\cite{Askari:ACC:2021}. 

\subsection{NMPC through the Lens of Bayesian Estimation}
\label{NMPC-Bayesian-view}

Consider a nonlinear dynamic system of the form: 
\begin{align}\label{state-equation}
x_{k+1} = f(x_{k},u_{k}),
\end{align}
where $x_{k} \in \mathbb{R}^{n_x}$ is the system state, and $u_{k} \in \mathbb{R}^{n_u}$ is the control input. The nonlinear mapping $f:  \mathbb{R}^{n_x} \times \mathbb{R}^{n_u} \rightarrow \mathbb{R}^{n_x}$ characterizes  the state transition. Here, it  represents  a feedforward neural network   model learned from data made on the actual system, or a hybrid model combining  physics with neural networks. 
 The system is subject to the following inequality constraints:
\begin{align}\label{inequality-constraints}
g_j({x}_k,u_{k})\leq 0,  \ \forall j=1,\ldots,m,
\end{align}
where $m$ is the total number of constraints.  We consider an   NMPC problem  such that $x_k$ tracks a reference signal $r_k$, which is stated as follows:
\begin{subequations} \label{NMPC-Standard}
\begin{align}
\min_{u_{k:k+H}} \quad &  \sum_{t=k}^{k+H}
  u_t^\top  Q u_t + ( x_t -  r_t)^\top R  ( x_t -  r_t),\\
\mathrm{s.t.} \quad & x_{t+1} = f(x_{t},u_{t}) , \\ \label{ineq-constraint}
& g_j(x_t,u_t) \leq 0 \quad \forall j=1,\ldots,m,\\ \nonumber
& t = k, \ldots, k+H,
\end{align}
\end{subequations}
where $H$ is the length of the upcoming horizon,  $u_{k:k+H} = \left\{u_k, u_{k+1}, \ldots, u_{k+H}\right\}$, and  $Q$ and $R$  are weighting matrices.  The above problem is solved   through time in a receding-horizon manner by computing   the optimal control input sequence $u_{k:k+H}^\ast$.  At every time, the first element, $u_k^\ast$  will be applied to control the system, and the rest  discarded. The same   optimization and control procedure will repeat itself recursively at the future time instants.  

While the NMPC problem in~\eqref{NMPC-Standard} is usually solved through numerical optimization in the literature, it can be interpreted as a Bayesian  estimation problem and thus addressed.  The following theorem shows this connection~\cite{Askari:ACC:2021,Stahl:SCL:2011}. 
\begin{thm}\label{ML-NMPC-Equivalence}
For the horizon $t=k,\ldots,k+H$, consider the virtual system 
\begin{align}\label{Virtual-System}
\left\{
\begin{aligned}
x_{t+1} &= f(x_t, u_t),\\
u_{t+1} &=  w_t,\\
r_t &= x_t+ v_t,
\end{aligned}
\right. 
\end{align}
where   $w_t \sim \mathcal{N} \left(0,Q^{-1} \right)$, $v_t \sim \mathcal{N} \left(0, R^{-1} \right)$, and the reference $r_t$ serves as virtual measurements. The maximum likelihood estimation of $x_t$ and $u_t$ via
\begin{align}\label{ML-Formulation}
\max_{x_{k:k+H}, u_{k:k+H}} \log \p(x_{k:k+H}, u_{k:k+H}   |  r_{k:k+H}),
\end{align}
where $x_{k:k+H} = \left\{x_k,x_{k+1}, \ldots, x_{k+H}\right\}$ (similarly for $u_{k:k+H} $ and $r_{k:k+H} $),  is equivalent to the NMPC problem in~\eqref{NMPC-Standard} without the inequality constraints in~\eqref{ineq-constraint}.
\end{thm}

Theorem~\ref{ML-NMPC-Equivalence} indicates that the original   NMPC problem can be converted into the problem of {\em estimating} the optimal control actions along with states given the specified reference. This viewpoint then ushers a new way of treating NMPC through estimation methods.

Before proceeding further, we explain how to incorporate the inequality constraints into a Bayesian estimation procedure, since they are an integral part of the NMPC formulation. Here, we adopt the barrier function method   to create  virtual measurements about the constraint satisfaction: 
\begin{align}\label{Virtual-measurement-barrier-function}
z_t = \phi\left(g( {x}_t, u_t)\right) + \eta_t,
\end{align}
where $z$ is the virtual measurement variable, $g$ is the collection of $g_j$ for $j=1,\ldots, m$, $\eta$ is an additive small noise, and $\phi$ is a barrier function.
Here, $\phi$ is chosen to be the softplus function:
\begin{equation}
 \phi{\left( s \right)} = \frac{1}{\alpha}\ln\left(1+\exp(\beta s )\right),
\end{equation}
which is parametrized by two tuning factors $\alpha$ and $\beta$.
Note that $\phi$ is  fully continuous  and   through appropriate parameterization, outputs almost zero at a point within the constraint set and   large values at points outside the set.  Hence,  $\phi$ can be used to quantify the constraint satisfaction or violation, and we let the virtual measurement $z_t$  be $0$. Further, we can include $z_t$ into~\eqref{Virtual-System} to allow estimation with an awareness of constraints. 

Now, consider~\eqref{Virtual-System} along with $z_t$ and rewrite it compactly in an augmented form:
\begin{align}\label{Virtual-System-Compact}
\left\{
\begin{aligned}
\bar{x}_{t+1} &= \bar{f}(\bar{x}_t) + \bar{w}_t, \\
\bar r_t &=\bar h(  \bar{x}_t ) +\bar v_t,
\end{aligned}
\right.
\end{align}
for $t = k, \ldots, k+H$, 
where $\bar{x}_t=\left[x_{t}^{\top} \ u_{t}^{\top}\right]^{\top}$, $\bar{w}_t=\left[0^\top \ w_t^{\top}\right]^{\top}$,  $\bar{v}_t=\left[v_t^\top \ \eta_t^{\top}\right]^{\top}$, $\bar f $ stems from $f$, and  $\bar h$ results from $r_t$ and $z_t$.  Based on the above, to address the original NMPC problem in~\eqref{NMPC-Standard}, we only need to perform state estimation for the above augmented system via
\begin{align} \label{Bayesian-Estimation-Augmented-System}
\max_{\bar x_{k:k+H}} \log \p( \bar x_{k:k+H}  |  \bar r_{k:k+H}).
\end{align}
 This can be achieved by particle filtering/smoothing, which is known as an effective means of state estimation for even highly nonlinear systems.  

\subsection{Development of the {CAP-NMPC} Approach} \label{CAP-NMPC-deployment}

The state estimation problem in~\eqref{Bayesian-Estimation-Augmented-System} involves both filtering and smoothing.
We first look at the forward filtering by considering $\p (\bar x_{k:t} | \bar r_{k:t})$ for $k \leq t \leq k+H$. Because it is practically impossible to obtain an analytical expression  of $\p (\bar x_{k:t} | \bar r_{k:t})$ for nonlinear systems, we  approximate it by using  a sample-based empirical distribution.  A common and useful approach is to do  importance sampling. That is, one draws samples from an alternative known distribution $\q (\bar x_{k:t} | \bar r_{k:t})$, which is called   importance or proposal distribution, and then evaluate  the weights of the samples in relation to $\p (\bar x_{k:t} | \bar r_{k:t})$. Suppose that $N$ samples, $\bar x_{k:t}^i$ for $i=1,\ldots,N$, are drawn from $\q (\bar x_{k:t} | \bar r_{k:t})$. Their importance weights are  given by   
\begin{align}\label{Weights}
W_t^i = \frac{\p (\bar x_{k:t} | \bar r_{k:t})}{\q (\bar x_{k:t} | \bar r_{k:t})},
\end{align}
which are then normalized to be between 0 and 1. As such, $\p (\bar x_{k:t} | \bar r_{k:t})$ can be  approximated as
\begin{align*}
\p (\bar x_{k:t} | \bar r_{k:t}) \approx \sum_{i=1}^N W_k^i \delta \left( \bar x_{k:t} - \bar x_{k:t}^i\right). 
\end{align*}
Note that~\eqref{Weights} also implies a recurrence relation in the  weight update:
\begin{align*}
W_t^i = \frac{\p (\bar x_{k:t} |\bar r_{k:t})}{\q (\bar x_{k:t} | \bar  r_{k:t})} = \frac{\p ( \bar r_{t} | \bar x_t^i) \p(\bar x_t^i | \bar x_{t-1}^i)}{\q ( \bar x_t^i | \bar x_{t-1}^i, \bar r_{k:t} )} W_{t-1}^i. 
\end{align*} 

One has different ways to implement the above procedure, with the key lying in choosing the importance distribution $q$. A  straightforward choice is to let $\q ( \bar x_t | \bar x_{t-1}, \bar r_{k:t} ) =  \p (\bar x_t | \bar x_{t-1})$. Given this choice,  we can draw samples  $\bar x_t^i \sim \p(\bar x_t | \bar x_{t-1}^i)$ at time $t$ and compute the associated normalized weights via
\begin{align}\label{const_weight}
W_t^i = \frac{\p(\bar r_t | \bar x_t^i)}{\sum_{j=1}^N \p(\bar r_t | \bar x_t^j)}.
\end{align}
This implementation is called  the {\em bootstrap particle filter}. For a particle filtering run,  a majority of the particles may have zero or almost zero weights after a few time steps. This is known as the issue of particle degeneracy, which lowers the overall quality of the particles   and reduces the estimation performance.  To resolve this issue, resampling can be used to replace  low-weight particles by those with high weights \cite{Sarkka:CUP:2013}.  

The backward smoothing follows the forward filtering as we only require $\p(\bar x_k | \bar r_{k:k+H})$, which can lead to a more accurate estimation of $\bar x_k$. There are different particle smoothers, and  the reweighting particle smoother will suffice here. It reweights the samples in a recursive  backward manner via
\begin{align}\label{reweighting}
W_{t|k+H}^i = \sum_{j=1}^N W_{t+1|k+H}^i\frac{W_t^i \p(\bar{x}_{t+1}^j|\bar{x}_t^i)}{ \sum_{l=1}^N W_t^l \p(\bar{x}_{t+1}^j|\bar{x}_t^l)},
\end{align}
where   $W_{k+H|k+H}^i = W_{k+H}^i$. The resultant procedure is called  {\em reweighted particle smoother}. After the smoothing,  the   empirical distribution for $\p(\bar{x}_k| \bar r_{k:k+H})$ is given by
\begin{equation*}
\p(\bar{x}_k| \bar r_{k:k+H})\approx\sum_{i=1}^N W_{k|k+H}^i\delta(\bar{x}_k-\bar{x}_k^i).
\end{equation*}
Finally, the best estimate of $\bar x_k$ from $\bar r_{k:k+H}$ is
\begin{align}\label{estimate}
\hat{\bar x}_k^\ast =  \mathbb{E}\left(\bar x_k \, | \, \bar r_{k:k+H}\right) = \sum_{i=1}^N W_{k|k+H}^i \bar x_k^i,
\end{align}
from which   the optimal control input  $u_k^*$  can be read.

Summarizing the above, the CAP-NMPC approach can be outlined as in Algorithm~\ref{NMPC-CA-PF/S}. It presents a fully sampling-based implementation of NMPC, which is promising for control of neural network dynamics. It should also be noted that this approach can well admit other realizations of particle filtering/smoothing, depending on the needs for sophistication and accuracy of estimation.

\begin{algorithm}[t]
\fontsize{10}{10}
  \caption{CAP-NMPC: NMPC via Constraint-Aware Particle Filtering/Smoothing}	  \label{NMPC-CA-PF/S}
  \begin{algorithmic}[1]
\State {Set up   NMPC}   by specifying the dynamic system~\eqref{state-equation}

\State {Recast NMPC} as particle filtering/smoothing by setting up the virtual system~\eqref{Virtual-System-Compact}

\For{$k=1, \ldots, T$}

\item[]{\hspace{14pt}\bf\em    Forward filtering}
 
\For{$t = k,\ldots, k+H$}

\If{$t=k$}

\State Draw samples $\bar x_k^i \sim  p(\bar x_k)$, $i=1, \ldots, N$

\Else

\State Draw samples $\bar x_t^i \sim \p(\bar x_t | \bar x_{t-1}^i)$, $i=1,\ldots,N$

\State Evaluate sample weights via~\eqref{const_weight}

\State Do resampling based on the weights

\EndIf

\EndFor

\item[]{\hspace{14pt}\bf\em    Backward smoothing}

\For{$t = k+H,\ldots, k$}

\If{$t=k+H$}

\State Assign $W_{k+H|k+H}^i = W_{k+H}^i$, $i=1,\ldots,N$

\Else 

\State Reweight the particles via~\eqref{reweighting}

\EndIf

\EndFor
 
\State Compute the optimal estimation of $\bar x_k$ via~\eqref{reweighting}

\State Export $u_k^\ast$ from $\bar x_k^\ast$, and apply it to the system~\eqref{state-equation}

\EndFor

\end{algorithmic}
\end{algorithm}

\section{Motion Planning Using Neural Network Dynamics \textit{via} CAP-NMPC} \label{motion-planning-formulation} 

In this section, we present the autonomous vehicle motion planning problem. 

\subsection{Motion Planning Problem Formulation}

We consider an autonomous vehicle whose goal is to arrive at the desired goal state $x^g$ from an initial state $x_0$   while considering its own dynamics~\eqref{state-equation} and constraints imposed by the surrounding~\cite{Chen:TIV:2019}. The objective of planning is to find an optimal trajectory without violating the constraints. To this end, an NMPC problem can be formulated as
\begin{subequations} \label{Motion-planning-NMPC}
\begin{align}
\min_{u_{k:k+H}} \quad &  \sum_{t=k}^{k+H}
  u_t^\top  Q u_t + ( x_t -  x^g)^\top R  ( x_t -  x^g) ,\\ \label{motion-planning-model}
\mathrm{s.t.} \quad & x_{t+1} = f(x_{t},u_{t}) , \\ \label{motion-planning-inequality-constraints}
& g_j(x_t,u_t) \leq 0 \quad \forall j=1,\ldots,m,\\ \nonumber
& t = k, \ldots, k+H,
\end{align}
\end{subequations}
which follows~\eqref{NMPC-Standard} by assigning the goal state to be the reference.  The model in~\eqref{motion-planning-model} captures the vehicle dynamics, and~\eqref{motion-planning-inequality-constraints} encompasses all the constraints due to vehicle operation or planning scenarios.

A mission of motion planning   is to make the autonomous vehicle  perform as well as or even better, than human drivers in a wide array of conditions or situations, and a key is using capable models. Despite the utility, physics-based  vehicle models have practical limitations, which may contain errors or biases, or suit only certain specific conditions. Meanwhile, the increasing abundance of data generated by autonomous vehicles makes it possible to construct precise, broadly applicable data-driven models. Among them, neural networks have shown tremendous merits for model-based control of vehicles~\cite{Spielberg:SR:2019}. Therefore, we will leverage a neural network model for the considered motion planning problem, with more details offered in~\ref{neural-network-dynamics}. Further,  a sampling-based implementation of  NMPC better suits control of neural networks than numerical optimization, as argued in Sections~\ref{introduction}-\ref{literature}. Recent studies also suggest sampling can be more competitive than numerical optimization in dealing with large-scale problems~\cite{Ma-Chen-Flammarion-Jordan:PNAS:2019}. We hence will use the CAP-NMPC approach to address the motion planning problem~\eqref{Motion-planning-NMPC} with neural network dynamics.

\subsection{The Neural Network Model}
\label{neural-network-dynamics}

A feedforward neural network is utilized to represent the vehicle dynamics. We describe the neural-network-based parameterization of  the vehicle system  and then explain the data collection and training process. 

\subsubsection{Vehicle Dynamics Parameterization}\label{sec:dynamic-system-parameterization} 
There are different approaches to parameterize the vehicle dynamics using neural networks.  A simple approach is to feed the state $x_{k}$ and control $u_{k}$ to a neural network  and make the network predict the next state of the system $x_{k+1}$. However, learning such a function could be difficult due to the small sampling time $\Delta t$. This leads to $x_k$ and $x_{k+1}$ being very similar in making, causing the  network to effectively learn an identity transformation~\cite{Nagabandi:ICRA:2018}.  Hence, we instead focus on parameterizing the state transition function, as it shows the incremental changes in the state. The  parametrization is then given by  
\begin{align*}
x_{k+1} = x_{k} + \Delta t\hat{f}(x_{k}, u_{k}; \theta),
\end{align*}
where $ \hat{f}$ represents the neural network to approximate the vehicle dynamics, and $\theta$ collects the weights of the network.

\subsubsection{Data collection and pre-processing}\label{sec:data-collection}

We generate and collect the training data from the considered vehicles model.
This is attained by sampling a set of states from a uniform distribution $x_{k} \sim \mathrm{Uni}(\mathcal{X}_k)$ and applying a random control, which is also sampled from a uniform distribution $u_{k} \sim \mathrm{Uni}(\mathcal{U}_k)$. Here, $\mathcal{X}$ and $\mathcal{U}$ are the respective feasible state and control constraint sets defined by the inequality constraints~\eqref{ineq-constraint}. Then, $x_k$ and $u_k$ are applied to the vehicle model, and the next state $x_{k+1}$ of the system is recorded. This constructs the required transition $\{x_{k}, u_{k}, x_{k+1}\}$ for training. The input to the neural network is the state and control pair $(x_{k}, u_{k})$, and the corresponding output is incremental state change $\Delta x_k=x_{k+1} - x_{k}$. The input and output are also normalized to ensure an almost equal contribution of each element of the state and control to the loss function. The obtained training dataset is $\mathcal{D}$ containing a million data points.

\subsubsection{Neural Network Structure and Training}\label{sec:training-nn} 

We use a dense neural network with four hidden layers of size $200, 300, 300, 100$, respectively, with the rectified linear function as the activation function. We then train the neural network by minimizing the mean squared error
\begin{align}
\mathcal{L}(\theta; \mathcal{D}) = \frac{1}{M} \sum_{i=1}^M  \norm{\Delta  x_{k}^{(i)} - \hat{f}(x_{k}^{(i)}, u_{k}^{(i)}; \theta)}^{2},
\label{eq:neural_net_loss}
\end{align}
where $M$ is the number of training samples in  $\mathcal{D}$. The training is done using Adam optimizer~\cite{Kingma:ICLR:2015}. We also use a different test dataset $\mathcal{D}_{\mathrm{test}}$ to evaluate the prediction accuracy of the trained neural network over data points not included in the training set. Finally, we point out that such a dense network is used in order to better validate the CAP-NMPC approach, even though a less dense one could also sufficiently approximate the vehicle dynamics. 

\begin{figure}[t]
	  \centering
	  \includegraphics[trim=25 80 35 100 ,clip,width=\linewidth]{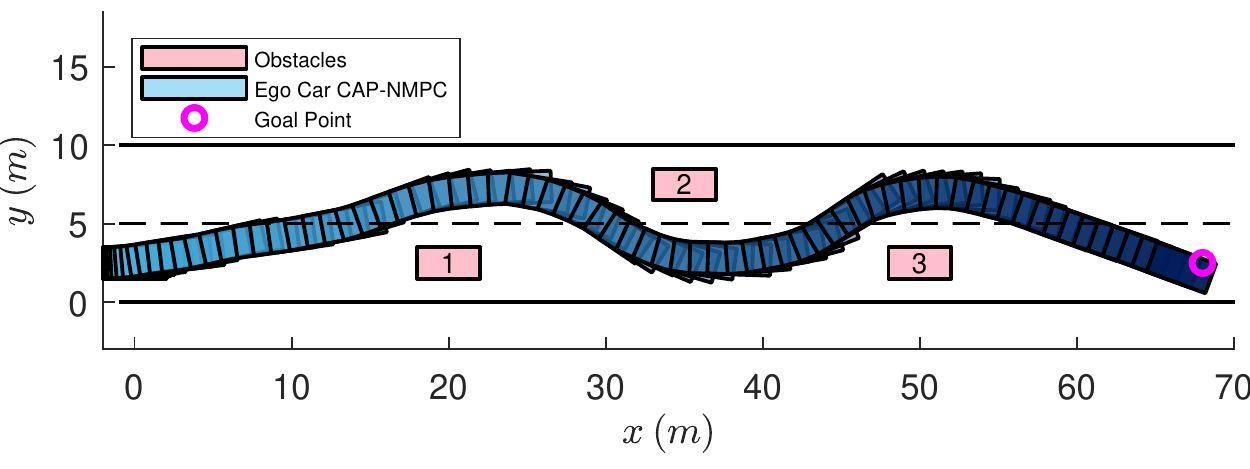}  
	  \caption{Planned trajectory for   Scenario 1 using the CAP-NMPC approach.}
	  \label{fig:straight-track-static-traj}
\end{figure}

\begin{figure}[t]
	  \centering
	  \includegraphics[trim = 25 15 30 15,clip,width=\linewidth]{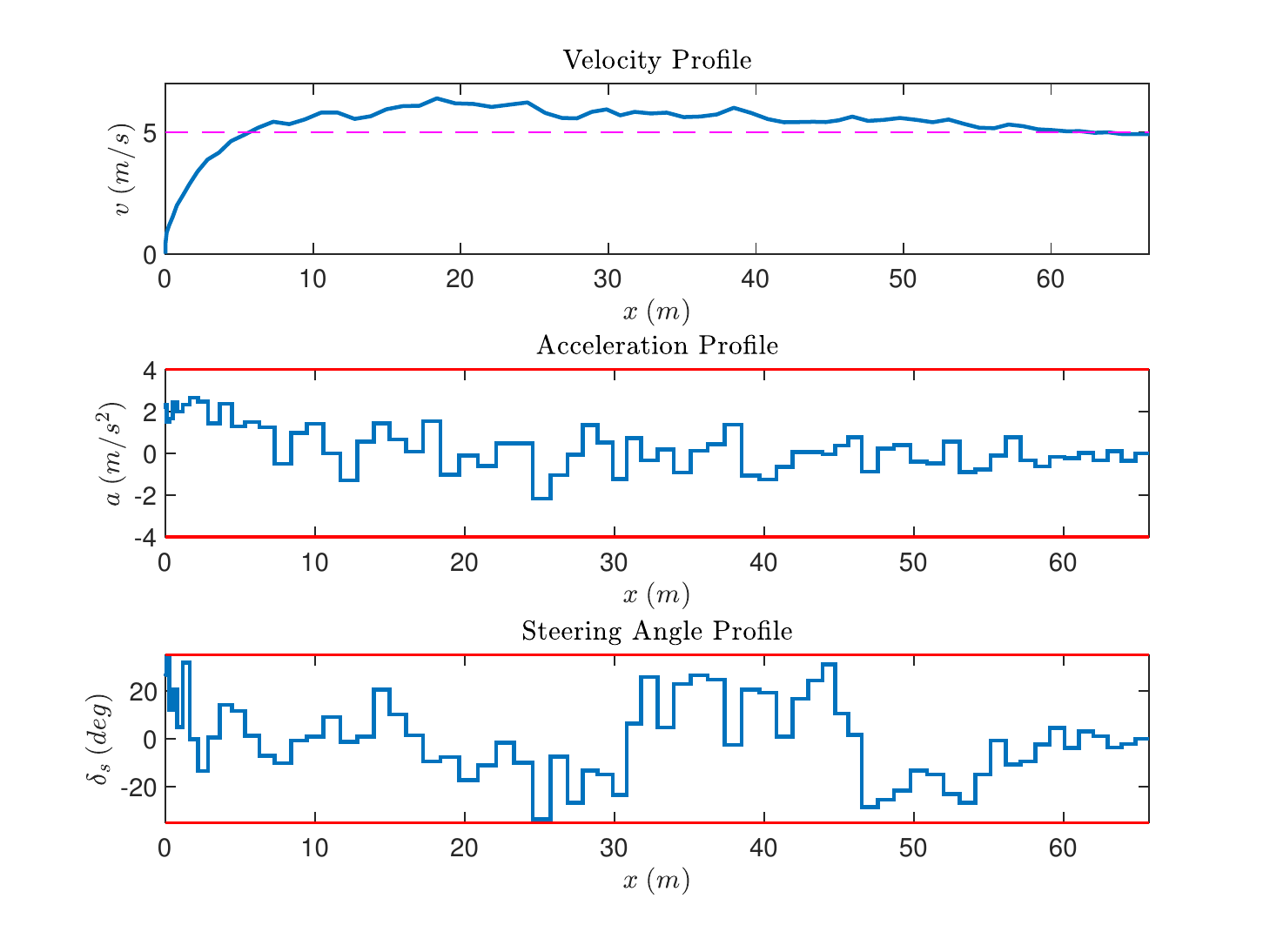}  
	  \caption{Velocity, acceleration and steering profiles for Scenario 1.}
	  \label{fig:straight-track-static-control}
\end{figure}
\section{Numerical Simulation}\label{numerical-sim}

Based on Section~\ref{motion-planning-formulation}, we now illustrate the effectiveness of the CAP-NMPC approach for motion planning.
 
\subsection{Constraints}\label{constraint-def}

The planning vehicle, or ego car, has a state $x_k$  including its 2D position $(x_k^p,y_k^p)$, linear velocity $\nu_k$, and heading angle $\psi_k$, as in the considered single-track vehicle model~\cite{rajmani}. Its control input  $u_k$ includes the acceleration $a_k$ and steering angle $\delta_k$. Three constraints are considered in the simulation study: road boundaries, obstacle collision avoidance, and the limits on the control actions.

\subsubsection{Road Boundary Constraints}\label{road-boundaries}
The  ego car is constrained to stay within the road boundaries. The maximum orthogonal distance $d_B$ from the road boundary is
\begin{align*} \label{distance-to-track}
d_B(x_k) = \max_j \left\{ \norm{x_k-B_j}_2 \right\},
\end{align*}
where $B_j$ denotes the orthogonal point on the $j$-th boundary of the road from the ego car. We require $d_B(x_k)\leq L$, where $L$ is the width of the road, such that  the ego car lies inside the road boundaries.

\subsubsection{Obstacle Avoidance Constraint} \label{obstacle-avoidance}The ego car can encounter many other vehicles in its path and should keep a safe distance from them. The distance between the ego car and the closest obstacles is expressed as
\begin{align*}
d_O(x_k) = \min_l  \norm{x_k-O_l}_2 ,
\end{align*}
where $O_l$ is the $l$-th obstacle. To ensure safety, we enforce  $d_O(x_k) \geq 0$.

\begin{figure}[t]
	  \centering
	  \includegraphics[trim=15 5 15 15,clip,width=\linewidth]{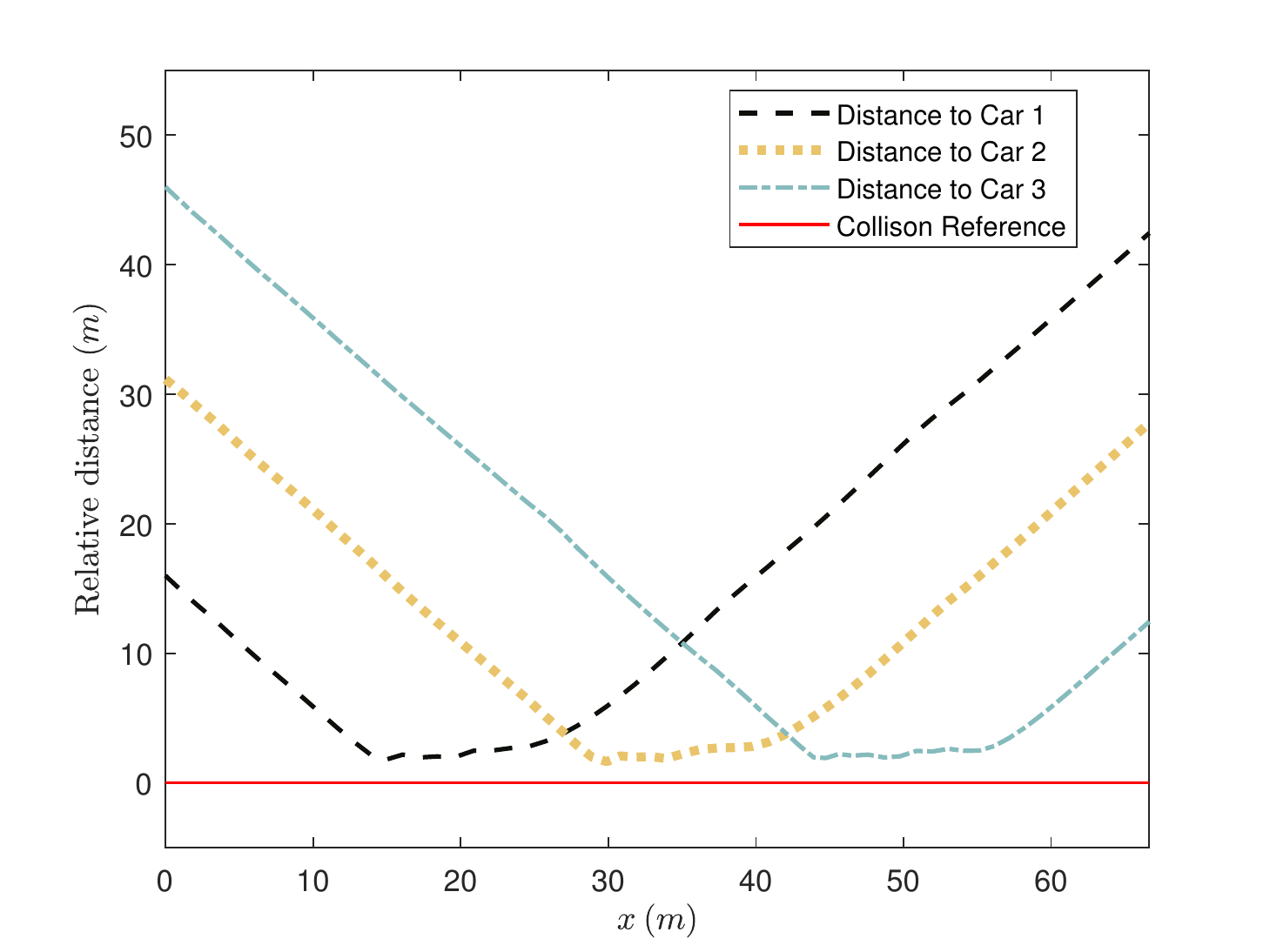}  
	  \caption{Minimum relative distances between the ego car and static obstacles for Scenario 1.}
	  \label{fig:sine-track-static-dist}
\end{figure}
\subsubsection{Control Constraints}
The maneuverability of the car is limited in real world. Here, we represent these limits by specifying the control  bounds:
\begin{equation*}\label{control-bounds}
\underline{u} \leq u_k \leq \overline{u},
\end{equation*}
where $\underline{u}$ is the lower bound, and $\overline{u}$ is the upper bound. 
\begin{figure}[t]
	  \centering
	  \includegraphics[trim = 20 70 40 55 ,clip,width=\linewidth]{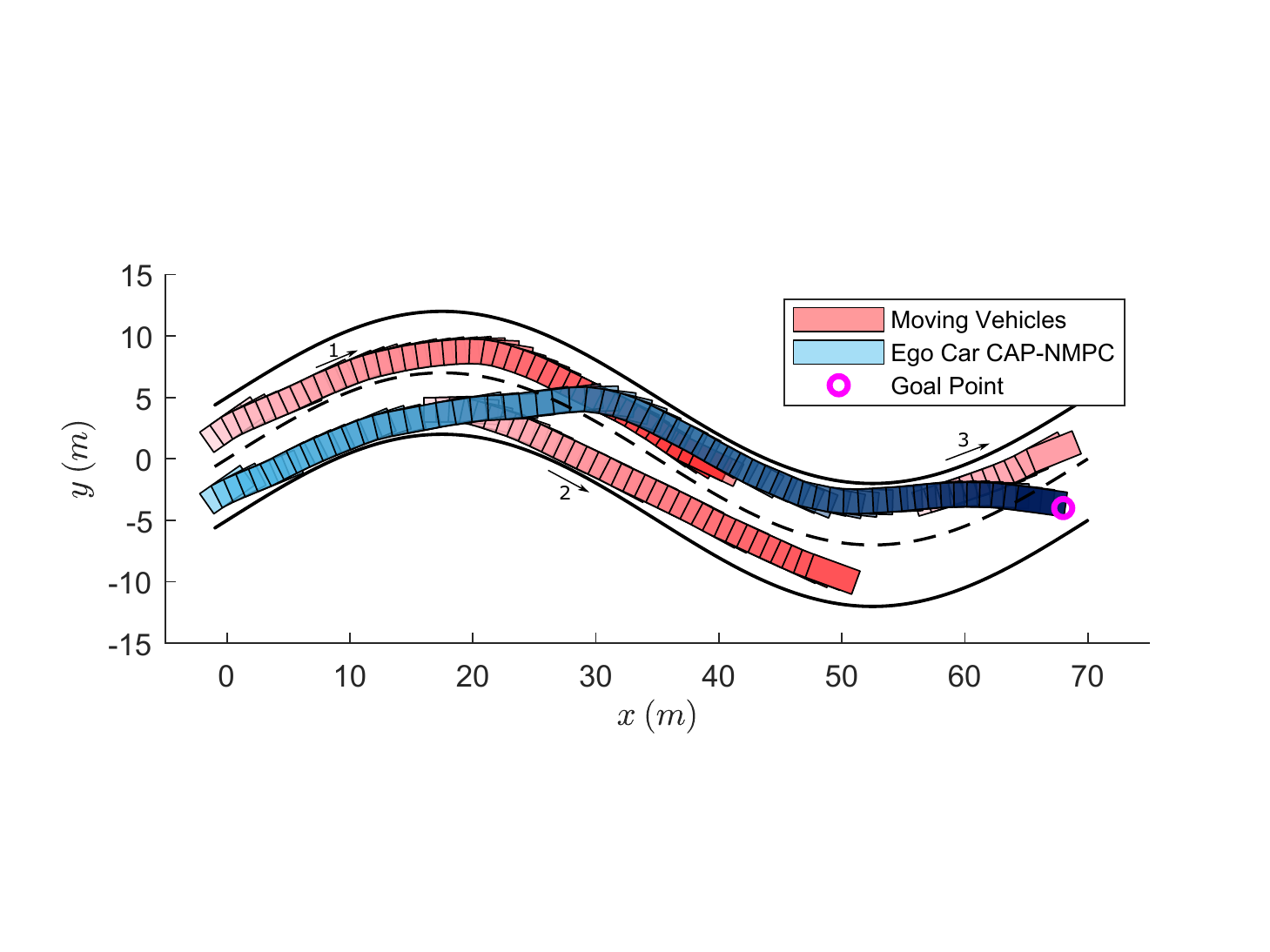}  
	  \caption{Planned trajectory for   Scenario 2 using the CAP-NMPC approach.}
	  \label{fig:sine-track-moving-traj}
\end{figure}
\begin{figure}[t]
	  \centering
	  \includegraphics[trim=25 15 30 15,clip,width=\linewidth]{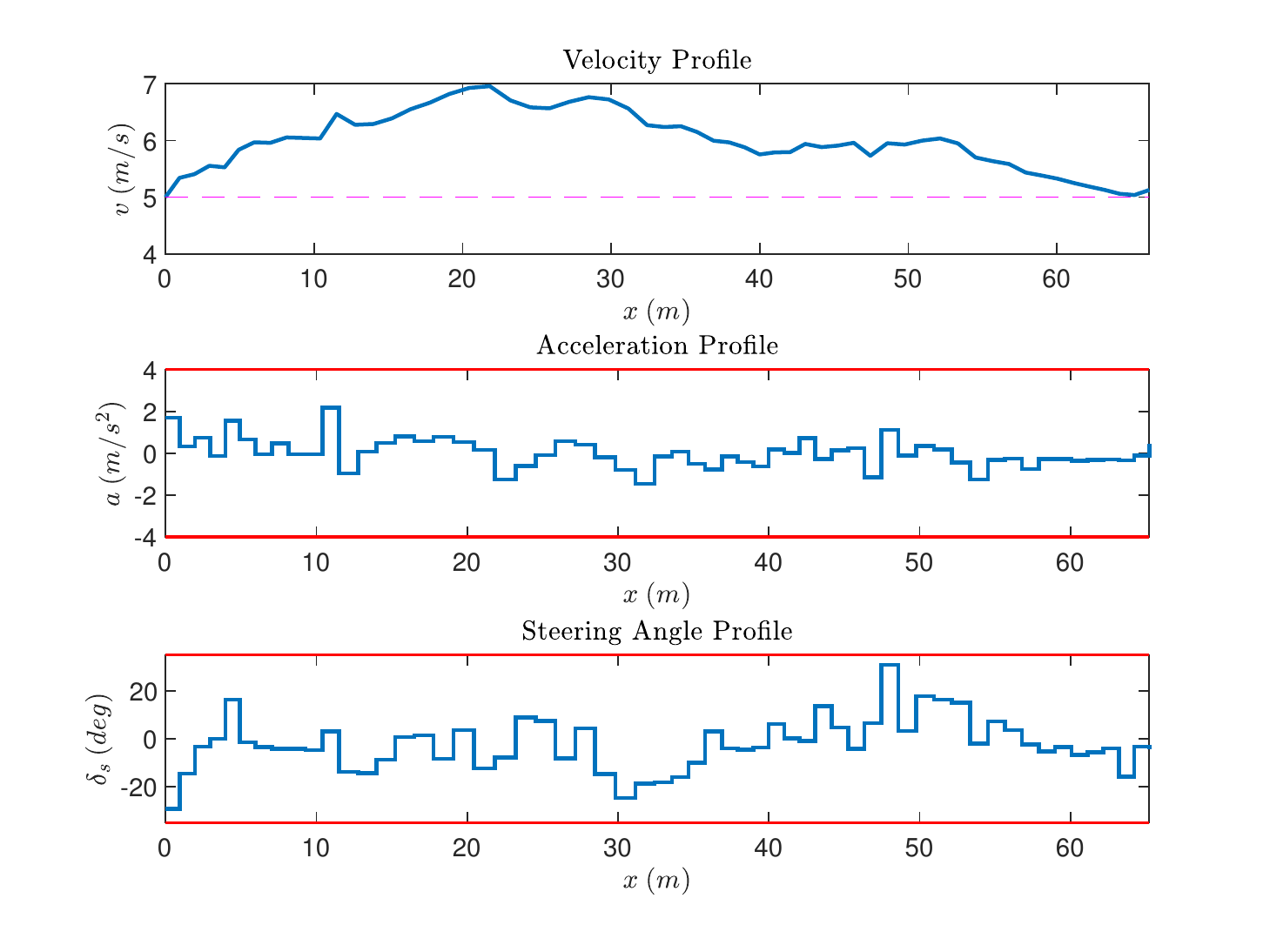}  
	  \caption{Velocity, acceleration and steering profiles for Scenario 2.}
	  \label{fig:sine-track-moving-control}
\end{figure}
\begin{figure}[t]
	  \centering
	  \includegraphics[trim=15 5 15 15,clip,width=\linewidth]{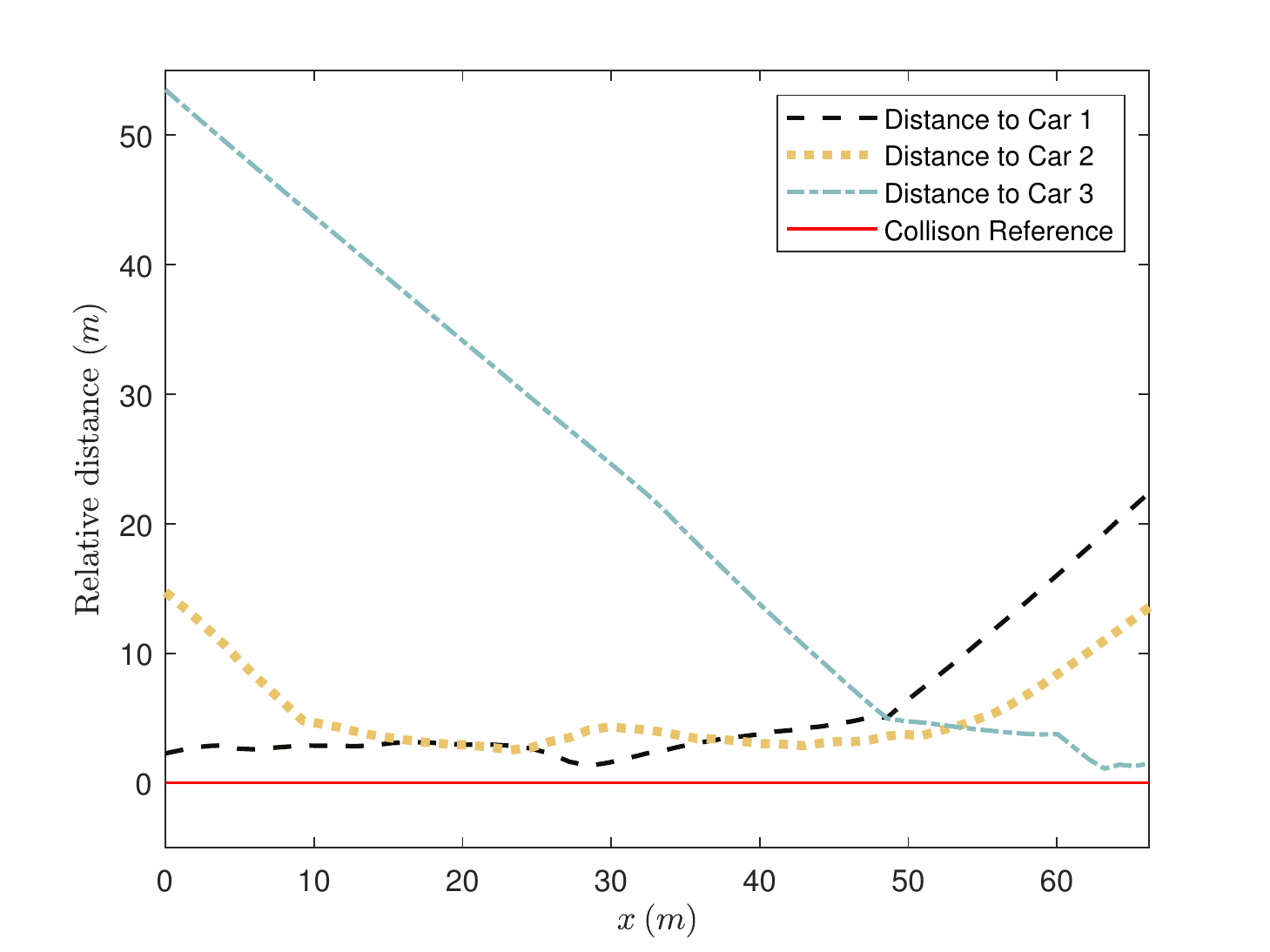}  
	  \caption{Minimum relative distances between the ego car and other moving vehicles in Scenario 2.}
	  \label{fig:sine-track-moving-dist}
\end{figure}

\subsection{Case Studies}
We consider two scenarios to examine the performance of the CAP-NMPC approach for   motion planning. In Scenario 1, the ego car is to travel on a straight road with static  obstacles. Scenario 2 involves  three moving vehicles on  a curved road and requires the ego car to overtake them. The prediction horizon length in the scenarios is $H=10$. The number of particles used in the CAP-NMPC is $N=300$. All the cars' lengths and widths are $4\: \mathrm{m}$ and $2\: \mathrm{m}$, respectively. The road width is $10\: \mathrm{m}$, and the sampling period $\Delta{t}=0.2\: \mathrm{s}$.

{\em Scenario 1: Straight road with static obstacles.} 
We place three stationary obstacles across the road. The ego car's trajectory is shown in Fig.~\ref{fig:straight-track-static-traj}, where the depth of the color from light to dark encodes the temporal information from present to future. The solid black line represents the road boundaries. We can observe that the ego vehicle is able to smoothly maneuver, bypassing the obstacles and keeping a safe distance at all times. Fig.~\ref{fig:straight-track-static-control} depicts the vehicles velocity and control profiles. It is evident that the CAP-NMPC allows the ego car to plan proficient maneuvers from zero velocity until reaching the goal state while satisfying the specified constraints and reference velocity (pink dotted line in Fig.~\ref{fig:straight-track-static-control}). It can be seen that the generated path with CAP-NMPC is indeed feasible for the vehicle to execute. In addition, in Fig.~\ref{fig:sine-track-static-dist}, we depict the minimum relative distance between the ego car and the obstacles, which is kept above zero to ensure collision avoidance.

{\em Scenario 2: Curved road with other moving vehicles.}
In this scenario, we consider a curved road with moving obstacles. The initial speed of all vehicles is $5 \:\mathrm{m/s}$ and the trajectories of the obstacle vehicles are obtained by assigning a goal position for each obstacle vehicle by CAP-NMPC algorithm. The obstacle vehicles are set in motion only when the ego car approaches them in order to ensure interaction with the ego car. This scenario is significantly more challenging given the dynamic environment, making collision avoidance a difficult task. For instance, observe from Figs. 4-6 the instant where the ego car is at $x = 20\:\mathrm{m}$. At this instant, the ego car has two other cars adjacent to it, one in the left lane and the other straight ahead. In addition, the road boundary is on the right of the ego car. The ego car has to carefully maneuver without leaving the road and colliding with the adjacent cars. To this end, the CAP-NMPC could successfully predict the upcoming situation and increase its velocity to smoothly pass between the two vehicles at a safe distance, as shown in Fig.~\ref{fig:sine-track-moving-dist}. Eventually, the ego car formulates a trajectory allowing it to overtake the other vehicles without collision to reach the specified goal state.

With this, we demonstrated the effectiveness of using a sample-based NMPC framework to solve the motion planning problem of a vehicle while considering neural network dynamics through the use of our proposed CAP-NMPC algorithm for realistic driving scenarios.

\section{Conclusion}

A growing  convergence of machine learning and advanced control is driving the frontiers of robotics and  stimulates new research problems. In this paper, we investigated the    problem of NMPC for neural network dynamics. The motivation lies in that, even though  NMPC has become an important method for robotics applications, its popular implementation based on numerical optimization can   meet only limited success when given neural network models. In a departure, we proposed to use a sampling-based NMPC approach, which built upon an Bayesian estimation perspective of NMPC and leveraged particle filtering/smoothing to estimate the best control actions. We then considered the motion planning problem for an autonomous vehicle with neural network dynamics and deployed the proposed approach to solve it under different driving conditions. The results demonstrated the potency of CAP-NMPC to handle constraints and compute   feasible trajectories effectively. 
Our future work will pursue integrating   the CAP-NMPC approach with other existing, very successful sampling-based motion planning techniques such as rapidly-exploring random trees or probabilistic roadmap methods, toward further improving the computational efficiency and feasibility in planning. The proposed approach can find prospective use in a wide variety of other robot control problems.

\balance
\bibliographystyle{IEEEtran}
\bibliography{PF-NMPC-NN-ICRA}

\begin{thebibliography}{10}
\providecommand{\url}[1]{#1}
\csname url@samestyle\endcsname
\providecommand{\newblock}{\relax}
\providecommand{\bibinfo}[2]{#2}
\providecommand{\BIBentrySTDinterwordspacing}{\spaceskip=0pt\relax}
\providecommand{\BIBentryALTinterwordstretchfactor}{4}
\providecommand{\BIBentryALTinterwordspacing}{\spaceskip=\fontdimen2\font plus
\BIBentryALTinterwordstretchfactor\fontdimen3\font minus
  \fontdimen4\font\relax}
\providecommand{\BIBforeignlanguage}[2]{{%
\expandafter\ifx\csname l@#1\endcsname\relax
\typeout{** WARNING: IEEEtran.bst: No hyphenation pattern has been}%
\typeout{** loaded for the language `#1'. Using the pattern for}%
\typeout{** the default language instead.}%
\else
\language=\csname l@#1\endcsname
\fi
#2}}
\providecommand{\BIBdecl}{\relax}
\BIBdecl

\bibitem{Nguyen-Tuong:CP:2011}
D.~Nguyen-Tuong and J.~Peters, ``Model learning for robot control: a survey,''
  \emph{Cognitive Processing}, vol.~12, no.~4, pp. 319--340, 2011.

\bibitem{Jordan:Science:2015}
M.~I. Jordan and T.~M. Mitchell, ``Machine learning: Trends, perspectives, and
  prospects,'' \emph{Science}, vol. 349, no. 6245, pp. 255--260, 2015.

\bibitem{Pierson:AR:2017}
H.~A. Pierson and M.~S. Gashler, ``Deep learning in robotics: a review of
  recent research,'' \emph{Advanced Robotics}, vol.~31, no.~16, pp. 821--835,
  2017.

\bibitem{Kober:IJRR:2013}
J.~Kober, J.~A. Bagnell, and J.~Peters, ``Reinforcement learning in robotics: A
  survey,'' \emph{The International Journal of Robotics Research}, vol.~32,
  no.~11, pp. 1238--1274, 2013.

\bibitem{Askari:ACC:2021}
I.~Askari, S.~Zeng, and H.~Fang, ``Nonlinear model predictive control based on
  constraint-aware particle filtering/smoothing,'' in \emph{2021 American
  Control Conference (ACC)}, 2021, pp. 3532--3537.

\bibitem{Draeger:CSM:1995}
A.~{Draeger}, S.~{Engell}, and H.~{Ranke}, ``Model predictive control using
  neural networks,'' \emph{IEEE Control Systems Magazine}, vol.~15, no.~5, pp.
  61--66, 1995.

\bibitem{Piche:NIPS:1999}
S.~Pich\'{e}, J.~Keeler, G.~Martin, G.~Boe, D.~Johnson, and M.~Gerules,
  ``Neural network based model predictive control,'' in \emph{Proceedings of
  the 12th International Conference on Neural Information Processing Systems},
  1999, p. 1029–1035.

\bibitem{Williams:ICRA:2017}
G.~{Williams}, N.~{Wagener}, B.~{Goldfain}, P.~{Drews}, J.~M. {Rehg},
  B.~{Boots}, and E.~A. {Theodorou}, ``Information theoretic \uppercase{MPC}
  for model-based reinforcement learning,'' in \emph{Proceedings of the IEEE
  International Conference on Robotics and Automation}, 2017, pp. 1714--1721.

\bibitem{Williams:TRO:2018}
G.~{Williams}, P.~{Drews}, B.~{Goldfain}, J.~M. {Rehg}, and E.~A. {Theodorou},
  ``Information-theoretic model predictive control: Theory and applications to
  autonomous driving,'' \emph{IEEE Transactions on Robotics}, vol.~34, no.~6,
  pp. 1603--1622, 2018.

\bibitem{Broad:arXiv:2018}
A.~Broad, I.~Abraham, T.~Murphey, and B.~Argall, ``Structured neural network
  dynamics for model-based control,'' \emph{ArXiv}, vol. abs/1808.01184, 2018.

\bibitem{Nagabandi:ICRA:2018}
A.~{Nagabandi}, G.~{Kahn}, R.~S. {Fearing}, and S.~{Levine}, ``Neural network
  dynamics for model-based deep reinforcement learning with model-free
  fine-tuning,'' in \emph{Proceedings of the IEEE International Conference on
  Robotics and Automation}, 2018, pp. 7559--7566.

\bibitem{Garimella:2018}
G.~Garimella and M.~Sheckells, ``Nonlinear model predictive control of an
  aerial manipulator using a recurrent neural network model,'' 2018.

\bibitem{Rankovic:IJCCC:2012}
V.~Rankovic, J.~Radulovic, N.~Grujovi\'{c}, and D.~Divac, ``Neural network
  model predictive control of nonlinear systems using genetic algorithms,''
  \emph{International Journal of Computers, Communications and Control},
  vol.~7, pp. 540--549, 2012.

\bibitem{Berntorp:TIV2019}
K.~{Berntorp}, T.~{Hoang}, and S.~{Di Cairano}, ``Motion planning of autonomous
  road vehicles by particle filtering,'' \emph{IEEE Transactions on Intelligent
  Vehicles}, vol.~4, no.~2, pp. 197--210, 2019.

\bibitem{Schwarting:ARCRAS:2018}
W.~Schwarting, J.~Alonso-Mora, and D.~Rus, ``Planning and decision-making for
  autonomous vehicles,'' \emph{Annual Review of Control, Robotics, and
  Autonomous Systems}, vol.~1, no.~1, pp. 187--210, 2018.

\bibitem{Gonzalez:TITS:2016}
D.~{Gonz\'{a}lez}, J.~{P\'{e}rez}, V.~{Milan\'{e}s}, and F.~{Nashashibi}, ``A
  review of motion planning techniques for automated vehicles,'' \emph{IEEE
  Transactions on Intelligent Transportation Systems}, vol.~17, no.~4, pp.
  1135--1145, 2016.

\bibitem{Paden:TIV:2016}
B.~{Paden}, M.~{Cap}, S.~Z. {Yong}, D.~{Yershov}, and E.~{Frazzoli}, ``A survey
  of motion planning and control techniques for self-driving urban vehicles,''
  \emph{IEEE Transactions on Intelligent Vehicles}, vol.~1, no.~1, pp. 33--55,
  2016.

\bibitem{Chen:TIV:2019}
J.~{Chen}, W.~{Zhan}, and M.~{Tomizuka}, ``Autonomous driving motion planning
  with constrained iterative {LQR},'' \emph{IEEE Transactions on Intelligent
  Vehicles}, vol.~4, no.~2, pp. 244--254, 2019.

\bibitem{Brunner:CDC:2017}
M.~{Brunner}, U.~{Rosolia}, J.~{Gonzales}, and F.~{Borrelli}, ``Repetitive
  learning model predictive control: An autonomous racing example,'' in
  \emph{Proceedings of the IEEE 56th Annual Conference on Decision and
  Control}, 2017, pp. 2545--2550.

\bibitem{Nolte:IVS:2017}
M.~{Nolte}, M.~{Rose}, T.~{Stolte}, and M.~{Maurer}, ``Model predictive control
  based trajectory generation for autonomous vehicles --- an architectural
  approach,'' in \emph{Proceedings of the IEEE Intelligent Vehicles Symposium},
  2017, pp. 798--805.

\bibitem{Cardoso:ICRA:2017}
V.~{Cardoso}, J.~{Oliveira}, T.~{Teixeira}, C.~{Badue}, F.~{Mutz},
  T.~{Oliveira-Santos}, L.~{Veronese}, and A.~F. {De Souza}, ``A
  model-predictive motion planner for the iara autonomous car,'' in
  \emph{Proceedings of the IEEE International Conference on Robotics and
  Automation}, 2017, pp. 225--230.

\bibitem{Liu:IVS:2017}
C.~{Liu}, S.~{Lee}, S.~{Varnhagen}, and H.~E. {Tseng}, ``Path planning for
  autonomous vehicles using model predictive control,'' in \emph{Proceedings of
  the IEEE Intelligent Vehicles Symposium (IV)}, 2017, pp. 174--179.

\bibitem{Gao:DSCC:2010}
Y.~Gao, T.~Lin, F.~Borrelli, E.~Tseng, and D.~Hrovat, ``Predictive control of
  autonomous ground vehicles with obstacle avoidance on slippery roads,'' in
  \emph{Proceedings of the ASME Dynamic Systems and Control Conference}, 2010,
  pp. 265--272.

\bibitem{Spielberg:SR:2019}
N.~A. Spielberg, M.~Brown, N.~R. Kapania, J.~C. Kegelman, and J.~C. Gerdes,
  ``Neural network vehicle models for high-performance automated driving,''
  \emph{Science Robotics}, vol.~4, no.~28, 2019.

\bibitem{Stahl:SCL:2011}
D.~Stahl and J.~Hauth, ``\uppercase{PF-MPC}: Particle filter-model predictive
  control,'' \emph{Systems \& Control Letters}, vol.~60, no.~8, pp. 632--643,
  2011.

\bibitem{Sarkka:CUP:2013}
S.~Särkkä, \emph{Bayesian Filtering and Smoothing}.\hskip 1em plus 0.5em
  minus 0.4em\relax Cambridge University Press, 2013.

\bibitem{Ma-Chen-Flammarion-Jordan:PNAS:2019}
Y.-A. Ma, Y.~Chen, C.~Jin, N.~Flammarion, and M.~I. Jordan, ``Sampling can be
  faster than optimization,'' \emph{Proceedings of the National Academy of
  Sciences}, vol. 116, no.~42, pp. 20\,881--20\,885, 2019.

\bibitem{Kingma:ICLR:2015}
D.~P. Kingma and J.~Ba, ``Adam: A method for stochastic optimization,'' in
  \emph{Proceedings of the International Conference on Learning
  Representations}, 2015.

\bibitem{rajmani}
R.~Rajamani, \emph{Vehicle Dynamics and Control}.\hskip 1em plus 0.5em minus
  0.4em\relax Springer US, 2012.

\end{thebibliography}
\end{document}